\documentclass[11pt,a4paper]{article}

\usepackage[utf8]{inputenc}
\usepackage[T1]{fontenc}
\usepackage{amsmath,amssymb,amsfonts}
\usepackage{graphicx}
\usepackage{booktabs}
\usepackage{hyperref}
\usepackage{xcolor}
\usepackage{geometry}
\usepackage{natbib}
\usepackage{microtype}
\usepackage{float}
\usepackage{enumitem}
\usepackage{caption}

\hypersetup{
  colorlinks=true,
  linkcolor=blue!70!black,
  citecolor=green!50!black,
  urlcolor=blue!70!black,
}

\title{%
  Autoregressive vs.\ Masked Diffusion Language Models:\\
  A Controlled Comparison%
}
\author{
  Caio Vicentino \\
  Independent Researcher \\
  \texttt{github.com/caiovicentino/arche}
}
\date{}

\begin{document}
\maketitle

% ── Abstract ────────────────────────────────────────────────────────────
\begin{abstract}
We present a controlled empirical comparison between autoregressive (AR) and masked diffusion (MDLM) language models.
Both models are trained on identical data (50M tokens from TinyStories), identical compute budget (20{,}000 steps, batch size 32, sequence length 512), and identical hardware (NVIDIA H100 80GB), isolating the generation paradigm as the sole variable.
We report three findings.
First, both paradigms achieve comparable training throughput (${\sim}$50K tokens/second), with MDLM requiring only 4.7\% more wall-clock time, countering the perception that diffusion training is substantially more expensive.
Second, AR converges faster and begins overfitting by step 14{,}000, while MDLM converges more slowly and is still improving at step 20{,}000, suggesting different compute-optimal training regimes.
Third, quantitative diversity analysis over 1{,}000 generated samples reveals a structural diversity--fluency trade-off: AR produces fluent but repetitive outputs (99.8\% begin with the same word), while MDLM generates more diverse narratives (93.4\% unique 5-word openings, higher Distinct-$n$, lower Self-BLEU), at the cost of occasional grammatical inconsistencies.
We note that validation losses across paradigms are not directly comparable, as they measure different objectives (next-token prediction vs.\ masked token prediction).
All code, trained checkpoints, and data pipelines are released for reproducibility.
\end{abstract}

% ── 1. Introduction ────────────────────────────────────────────────────
\section{Introduction}

Language models have been overwhelmingly autoregressive: tokens are generated left-to-right, each conditioned on all preceding tokens.
This paradigm underpins GPT \citep{radford2019gpt2}, LLaMA \citep{touvron2023llama}, and virtually all production language models.
Its strengths are well understood---efficient training via teacher forcing, straightforward sampling, and strong perplexity---but so are its limitations: sequential generation that cannot revise earlier tokens, and a strong left-to-right bias that can produce repetitive outputs when the most probable prefix dominates.

A fundamentally different paradigm has recently emerged: diffusion-based language models.
Adapted from the continuous diffusion framework that revolutionized image generation \citep{ho2020ddpm,song2021score}, these models generate text by starting from a fully corrupted (masked) sequence and iteratively denoising it.
Masked Diffusion Language Models (MDLM; \citealt{mdlm}) use a learned masking schedule to corrupt text during training and reverse the process during generation.
SEDD \citep{sedd} achieved ICML 2024 Best Paper with substantially improved generative perplexity over GPT-2.
LLaDA \citep{llada} demonstrated competitive performance at 8B parameters, matching LLaMA3 on downstream benchmarks.
Gemini Diffusion \citep{geminidiffusion} demonstrated fast generation at commercial quality.

Despite this rapid progress, controlled head-to-head comparisons remain scarce.
Existing studies either compare at different scales, use different datasets, or employ different compute budgets, making it difficult to attribute observed differences to the generation paradigm itself rather than confounding variables.

\paragraph{This work.}
We train an autoregressive Transformer and a masked diffusion Transformer on \emph{exactly the same} data, compute budget, and hardware.
The only differences are those inherent to each paradigm: causal vs.\ bidirectional attention, next-token vs.\ masked-token objective, and the MDLM's timestep conditioning module.
This controlled setup allows us to isolate the effect of the generation paradigm on training dynamics, convergence behavior, and generation quality.

We acknowledge upfront that our experiments are at small scale (123--163M parameters, 50M tokens) and that our generation quality analysis is qualitative rather than quantitative.
Our goal is not to declare a winner, but to provide a reproducible baseline comparison that highlights the distinct trade-offs of each paradigm.

\subsection{Contributions}

\begin{enumerate}[leftmargin=*,itemsep=2pt]
  \item \textbf{Controlled experimental setup}: AR and MDLM trained with identical data, steps, batch size, sequence length, model dimensions, optimizer, and hardware (Table~\ref{tab:setup}). To our knowledge, this is the most tightly controlled small-scale comparison of these paradigms to date.

  \item \textbf{Training dynamics analysis}: We provide step-by-step convergence curves (Table~\ref{tab:convergence}) showing that AR converges faster but overfits earlier, while MDLM converges slowly but steadily, suggesting different compute-optimal training lengths.

  \item \textbf{Throughput parity}: We find that MDLM training is only 4.7\% slower than AR in wall-clock time (113.0 vs.\ 107.9 minutes), countering the common assumption that diffusion training is substantially more expensive.

  \item \textbf{Quantitative diversity--fluency trade-off}: Over 1{,}000 generated samples, we show that MDLM achieves 93.4\% unique openings vs.\ AR's 3.3\%, with higher Distinct-$n$ and lower Self-BLEU, confirming a structural diversity advantage at the cost of fluency.

  \item \textbf{Full reproducibility}: All code, data pipelines, trained checkpoints (PyTorch), and evaluation scripts released at \url{https://github.com/caiovicentino/arche}.
\end{enumerate}

% ── 2. Related Work ────────────────────────────────────────────────────
\section{Related Work}

\subsection{Autoregressive Language Models}

The autoregressive Transformer \citep{vaswani2017attention,radford2019gpt2} remains the dominant language modeling paradigm.
Scaling laws \citep{kaplan2020scaling,hoffmann2022chinchilla} have established predictable relationships between model size, data, and performance.
Recent architectural innovations include Multi-Head Latent Attention for KV-cache compression \citep{deepseekv2}, Gated Linear Attention for efficient training \citep{yang2024gla}, and State Space Models for linear-time inference \citep{gu2023mamba}.
All of these operate within the autoregressive framework.

\subsection{Diffusion Language Models}

Adapting diffusion to discrete text has been an active area.
D3PM \citep{austin2021d3pm} introduced discrete diffusion with transition matrices.
MDLM \citep{mdlm} simplified this to a masking-based framework with a cosine schedule, achieving strong results with standard Transformer backbones.
SEDD \citep{sedd} used score-based estimation for discrete distributions, earning ICML 2024 Best Paper.
LLaDA \citep{llada} scaled masked diffusion to 8B parameters, demonstrating competitive downstream performance with LLaMA3.
Gemini Diffusion \citep{geminidiffusion} achieved commercial-grade generation with 5$\times$ speed improvements.

The key architectural difference from AR models is that diffusion models use \emph{bidirectional} attention (each position can attend to all unmasked positions) and condition on a noise level or timestep.
This bidirectional context is both a strength (richer conditioning) and a difference that complicates direct comparison.

\subsection{Existing Comparisons}

Prior comparisons between AR and diffusion language models exist but are not tightly controlled.
\citet{mdlm} compare against GPT-2 but at different parameter counts and with different training data.
\citet{llada} compare against LLaMA3 but with different training corpora and compute budgets.
\citet{sedd} report perplexity improvements over GPT-2 but using a different model architecture.
Our work fills the gap of a \emph{same-data, same-budget, same-hardware} comparison, albeit at small scale.

% ── 3. Method ──────────────────────────────────────────────────────────
\section{Experimental Setup}

\subsection{Controlled Variables}

Table~\ref{tab:setup} summarizes the experimental design.
All shared hyperparameters---data, training steps, batch size, sequence length, model dimensions, number of layers, number of heads, FFN dimension, optimizer, and hardware---are identical between the two models.

\begin{table}[t]
\centering
\caption{Controlled experimental setup. All shared variables are identical; differences arise solely from the paradigm.}
\label{tab:setup}
\smallskip
\begin{tabular}{@{}lcc@{}}
\toprule
Variable & AR & MDLM \\
\midrule
\multicolumn{3}{@{}l}{\textit{Shared (identical)}} \\
\quad Data & \multicolumn{2}{c}{TinyStories \citep{tinystories}, 50M tokens} \\
\quad Steps & \multicolumn{2}{c}{20{,}000} \\
\quad Batch size & \multicolumn{2}{c}{32} \\
\quad Sequence length & \multicolumn{2}{c}{512} \\
\quad $d_{\text{model}}$ & \multicolumn{2}{c}{768} \\
\quad Layers & \multicolumn{2}{c}{12} \\
\quad Heads & \multicolumn{2}{c}{12} \\
\quad FFN dim & \multicolumn{2}{c}{2{,}048} \\
\quad Optimizer & \multicolumn{2}{c}{AdamW (lr\,=\,3e-4, wd\,=\,0.01)} \\
\quad Hardware & \multicolumn{2}{c}{NVIDIA H100 80GB} \\
\midrule
\multicolumn{3}{@{}l}{\textit{Paradigm-specific (inherent differences)}} \\
\quad Attention mask & Causal & Bidirectional \\
\quad Training objective & Next-token prediction & Masked token prediction \\
\quad Noise schedule & N/A & Cosine \\
\quad Timestep conditioning & N/A & Sinusoidal embedding \\
\quad Parameters & 123.6M & 162.7M \\
\bottomrule
\end{tabular}
\end{table}

\subsection{Parameter Count Difference}
\label{sec:param-diff}

The MDLM model has 31.6\% more parameters than the AR model (162.7M vs.\ 123.6M).
This difference is not an oversight but reflects inherent architectural requirements:
\begin{itemize}[leftmargin=*,itemsep=2pt]
  \item \textbf{Bidirectional attention}: The MDLM uses full (non-causal) attention, which does not change the parameter count but does change the computation pattern.
  \item \textbf{Timestep conditioning}: The MDLM requires a timestep embedding module (sinusoidal embedding followed by projection layers) to condition on the noise level, adding parameters with no AR counterpart.
  \item \textbf{Output head}: The MDLM predicts tokens at all masked positions simultaneously, potentially requiring a slightly different output projection structure.
\end{itemize}
We chose not to artificially reduce the MDLM to match the AR parameter count, as this would compromise the MDLM design.
This difference should be considered when interpreting results: the MDLM has greater model capacity, which could favor it in principle, yet it converges more slowly in practice.

\subsection{Autoregressive Model}

The AR model is a standard decoder-only Transformer with causal (unidirectional) attention.
Each position attends only to itself and all preceding positions.
The training objective is standard next-token prediction: given a sequence $(x_1, \ldots, x_T)$, the model minimizes
\begin{equation}
  \mathcal{L}_{\text{AR}} = -\sum_{t=1}^{T} \log p_\theta(x_t \mid x_{<t})
\end{equation}
We use learned positional embeddings and pre-norm (LayerNorm before attention and FFN).

\subsection{Masked Diffusion Language Model (MDLM)}

The MDLM follows \citet{mdlm}.
The architecture is a bidirectional Transformer---identical to the AR model except that the causal mask is removed, allowing each position to attend to all other positions.
Additionally, a timestep embedding conditions the model on the current noise level.

\paragraph{Training.}
For each training example, we sample a random timestep $t \sim U(0,1)$ and mask tokens independently with probability
\begin{equation}
  \gamma(t) = 1 - \cos^2\!\left(\frac{\pi t}{2}\right) \quad \text{(cosine schedule)}
\end{equation}
Masked tokens are replaced with a special \texttt{[MASK]} token.
The model predicts the original token at each masked position, with cross-entropy loss computed only at masked positions:
\begin{equation}
  \mathcal{L}_{\text{MDLM}} = -\mathbb{E}_{t,\mathbf{m}} \sum_{i : m_i = 1} \log p_\theta(x_i \mid \tilde{\mathbf{x}}, t)
\end{equation}
where $\tilde{\mathbf{x}}$ is the masked input and $\mathbf{m}$ is the binary mask.

\paragraph{Generation.}
Starting from a fully masked sequence of length $L$, we iteratively unmask tokens over $S = 100$ steps.
At each step $s$, the model predicts all masked positions; we select the $\lfloor L / S \rfloor$ most confident predictions (highest predicted probability) and unmask them.
We apply temperature annealing (starting high for diversity, decreasing for coherence) and a repetition penalty to reduce degenerate outputs.

% ── 4. Results ─────────────────────────────────────────────────────────
\section{Results}

\subsection{Training Efficiency}

Both paradigms achieve comparable training throughput on the H100 (Table~\ref{tab:results}).
AR trains at 50{,}620 tokens/second; MDLM at 48{,}343 tokens/second (4.5\% slower).
Total wall-clock time differs by only 5.1 minutes (107.9 vs.\ 113.0 minutes for 20K steps).
This near-parity is notable: it suggests that the computational overhead of diffusion training (sampling timesteps, applying masks, computing loss only at masked positions) is modest relative to the dominant cost of the forward and backward passes through the Transformer.

\begin{table}[t]
\centering
\caption{AR vs.\ MDLM training comparison on H100 80GB (20K steps, TinyStories 50M tokens).}
\label{tab:results}
\smallskip
\begin{tabular}{@{}lcc@{}}
\toprule
Metric & AR & MDLM \\
\midrule
Parameters & 123.6M & 162.7M \\
Training time & 107.9 min & 113.0 min \\
Throughput & 50{,}620 tok/s & 48{,}343 tok/s \\
Best val loss & 1.589 & 3.412 \\
Best checkpoint & Step 14{,}000 & Step 20{,}000 \\
\bottomrule
\multicolumn{3}{@{}l}{\footnotesize Val losses use different objectives and are not cross-comparable.}
\end{tabular}
\end{table}

\subsection{Convergence Dynamics}

\begin{figure}[t]
\centering
\includegraphics[width=0.85\linewidth]{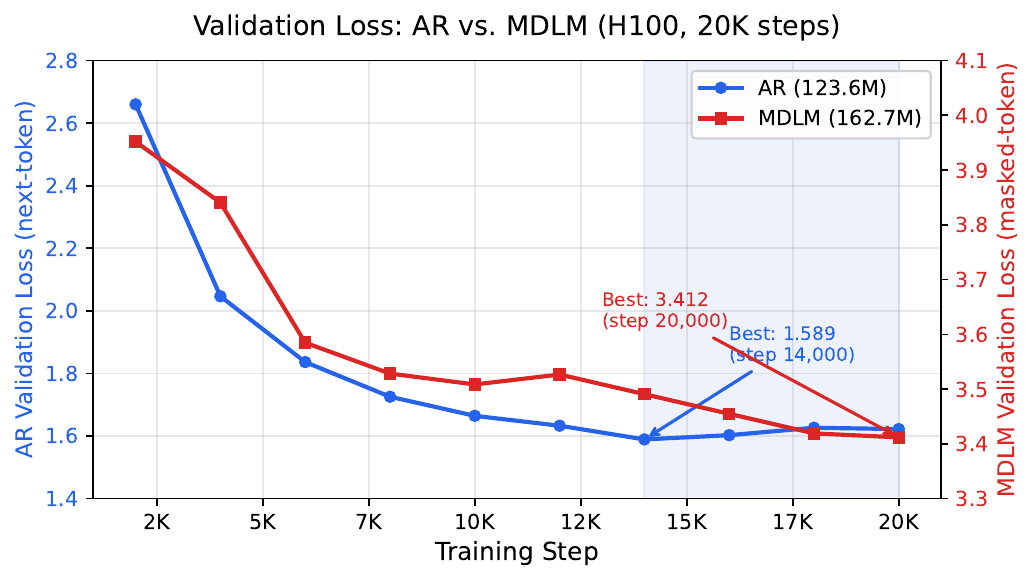}
\caption{Convergence comparison. Dual $y$-axes reflect different loss functions; the shape of each curve (not the absolute value) is the relevant comparison. AR overfits after step 14K (blue shaded region); MDLM continues improving monotonically.}
\label{fig:loss_curves}
\end{figure}

Table~\ref{tab:convergence} and Figure~\ref{fig:loss_curves} show validation loss over training.
The two loss functions measure different quantities---next-token prediction (AR) vs.\ masked token prediction with bidirectional context (MDLM)---so the absolute values are not cross-comparable.
However, the \emph{convergence patterns} are directly comparable and reveal a clear difference:

\begin{itemize}[leftmargin=*,itemsep=2pt]
  \item \textbf{AR converges fast, overfits early.}
  Validation loss reaches its minimum of 1.589 at step 14{,}000, then rises to 1.622 by step 20{,}000---a clear sign of overfitting on 50M tokens at this model size.

  \item \textbf{MDLM converges slowly, keeps improving.}
  Validation loss decreases monotonically throughout training, reaching 3.412 at step 20{,}000 with no sign of plateauing.
  This suggests MDLM would benefit from substantially longer training.
\end{itemize}

This asymmetry has practical implications: AR models at this scale need early stopping or regularization after ${\sim}$14K steps on 50M tokens, while MDLM models can productively train for longer.
Whether this reflects a fundamental property of the paradigm or simply the effect of the larger parameter count (Section~\ref{sec:param-diff}) is an open question.

\begin{table}[t]
\centering
\caption{Validation loss at selected checkpoints. Each column measures a different objective; read vertically for convergence trends, not horizontally.}
\label{tab:convergence}
\smallskip
\begin{tabular}{@{}rcc@{}}
\toprule
Step & AR Val Loss & MDLM Val Loss \\
\midrule
2{,}000 & 2.660 & 3.952 \\
4{,}000 & 2.047 & 3.841 \\
6{,}000 & 1.837 & 3.585 \\
8{,}000 & 1.726 & 3.528 \\
10{,}000 & 1.665 & 3.509 \\
12{,}000 & 1.633 & 3.527 \\
14{,}000 & \textbf{1.589} & 3.491 \\
16{,}000 & 1.603 & 3.455 \\
18{,}000 & 1.627 & 3.419 \\
20{,}000 & 1.622 & \textbf{3.412} \\
\bottomrule
\end{tabular}
\end{table}

\begin{figure}[t]
\centering
\includegraphics[width=0.85\linewidth]{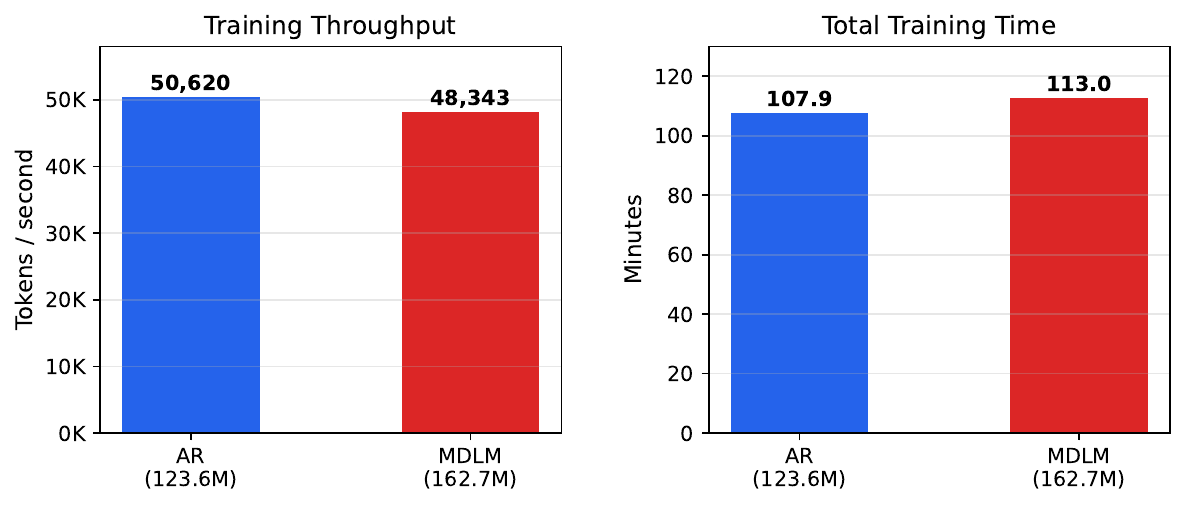}
\caption{Training throughput and wall-clock time. MDLM achieves 95.5\% of AR throughput, requiring only 5.1 additional minutes over 20K steps.}
\label{fig:throughput}
\end{figure}

\subsection{Generation Diversity: Quantitative Analysis}
\label{sec:generation}

We generate 1{,}000 independent samples from each model (128 tokens each) and compute standard diversity metrics.
AR uses nucleus sampling ($p = 0.9$, temperature 0.8); MDLM uses confidence-based unmasking (100 steps, temperature annealing 1.2$\to$0.5, repetition penalty 1.3).
Results are shown in Table~\ref{tab:diversity}.

\begin{table}[t]
\centering
\caption{Generation diversity metrics over 1{,}000 samples (128 tokens each). Distinct-$n$ measures the ratio of unique $n$-grams to total $n$-grams (higher = more diverse). Self-BLEU measures average BLEU of each sample against 50 random others (lower = more diverse). Unique first-5grams measures the fraction of samples with a unique 5-word opening.}
\label{tab:diversity}
\smallskip
\begin{tabular}{@{}lccc@{}}
\toprule
Metric & AR & MDLM & More diverse \\
\midrule
Distinct-1 & 0.053 & 0.058 & MDLM \\
Distinct-2 & 0.275 & 0.297 & MDLM \\
Distinct-3 & 0.531 & 0.542 & MDLM \\
Distinct-4 & 0.701 & 0.690 & AR \\
Self-BLEU ($\downarrow$) & 0.341 & 0.334 & MDLM \\
Vocabulary used & 5{,}360 & 5{,}754 & MDLM \\
Unique first words & 0.2\% & 36.1\% & MDLM \\
Unique first 5-grams & 3.3\% & 93.4\% & MDLM \\
\bottomrule
\end{tabular}
\end{table}

\paragraph{Token-level diversity is modest.}
Distinct-1 through Distinct-3 and Self-BLEU show MDLM as slightly more diverse, but the margins are small (e.g., Distinct-2: 0.297 vs.\ 0.275).
At the token and bigram level, both models draw from similar vocabulary when generating TinyStories-style text.
Distinct-4 slightly favors AR, likely because AR's longer coherent spans produce more unique 4-grams by construction.

\paragraph{Structural diversity is dramatic.}
The most striking result is in the opening statistics: 99.8\% of AR samples begin with the same word (``Once''), and only 3.3\% have a unique 5-word opening.
In contrast, 36.1\% of MDLM samples begin with a unique first word, and 93.4\% have a unique 5-word opening.
This quantifies the ``prefix mode collapse'' phenomenon: AR generation is dominated by the single most probable prefix in TinyStories (``Once upon a time, there was a\ldots''), while MDLM's non-sequential generation avoids this trap entirely.

\paragraph{Qualitative examples.}
To illustrate the quantitative findings, we show representative samples from each model:

\begin{quote}
\small
\textbf{AR-1}: ``Once upon a time, there was a mommy and a baby. They both wanted to go to the store. The mommy wanted to buy a new dress, but the baby was too small\ldots''

\textbf{AR-2}: ``Once upon a time, there was a little girl named Sarah. She was only three years old. One day, Sarah was walking in the park\ldots''
\end{quote}

\begin{quote}
\small
\textbf{MDLM-1}: ``Mom asks. `Yes, we can,' Anna and Ben say. They run to the door. Mom opens the door. She sees their faces! She hugs\ldots''

\textbf{MDLM-2}: ``\,`I am so glad you hugged me too.' She replied: `Yes -- I'm proud of you!' Once upon a time there were two best friends\ldots''
\end{quote}

AR samples are fluent and coherent but structurally identical; MDLM samples begin \emph{in medias res} with varied characters and situations, but exhibit occasional grammatical artifacts.

\paragraph{Interpretation.}
The structural diversity difference is architectural.
In AR generation, the model commits to tokens left-to-right: once the most probable prefix is generated (``Once upon a time''), the conditional distribution over remaining tokens is heavily constrained.
In MDLM generation, all positions start as \texttt{[MASK]} and are revealed based on confidence---the model can generate the middle of a narrative before the beginning, avoiding prefix mode collapse.

This parallels findings in image generation, where diffusion models produce higher-diversity outputs than autoregressive pixel models \citep{ho2020ddpm}. Our quantitative results confirm this trade-off extends to language.

% ── 5. Discussion ──────────────────────────────────────────────────────
\section{Discussion}

\subsection{Throughput Parity Is Surprising}

A common concern about diffusion language models is training cost.
Our finding that MDLM trains at 95.5\% of AR throughput (48{,}343 vs.\ 50{,}620 tokens/second) suggests that the per-step overhead of diffusion training---sampling timesteps, constructing masks, computing loss only at masked positions---is negligible relative to the Transformer forward/backward pass.
At larger scales where the Transformer computation dominates even more, the gap may shrink further.

This does \emph{not} mean the two paradigms have equal sample efficiency.
AR reached its best validation loss at 14K steps; MDLM was still improving at 20K steps.
If MDLM requires, say, 2--3$\times$ more steps to converge, the total training cost would be correspondingly higher despite similar per-step throughput.
Characterizing the step-efficiency gap is an important question for future work.

\subsection{Different Convergence Regimes}

The divergent overfitting behavior---AR overfits at 14K steps while MDLM continues improving at 20K---is consistent with a capacity-utilization hypothesis.
AR sees each token in a fixed left-to-right context; it can memorize common patterns quickly.
MDLM sees each token in a randomly masked context that varies every epoch; this implicit data augmentation may act as a regularizer, slowing convergence but reducing overfitting.

Alternatively, the difference may simply reflect the parameter count gap (Section~\ref{sec:param-diff}): the MDLM's larger model may need more steps to fit the data.
Disentangling these explanations requires experiments with matched parameter counts, which we leave for future work.

\subsection{The Diversity--Fluency Trade-off}

Our quantitative results (Table~\ref{tab:diversity}) confirm that AR and MDLM are \emph{complementary} rather than competing paradigms.
The trade-off manifests most clearly at the structural level: AR achieves perfect fluency but collapses to a single narrative template (99.8\% identical first word), while MDLM produces 93.4\% unique openings at the cost of occasional grammatical errors.
\begin{itemize}[leftmargin=*,itemsep=2pt]
  \item \textbf{AR} excels when fluency and coherence are paramount (translation, summarization, code generation).
  \item \textbf{MDLM} may be preferable when diversity is valued (creative writing, data augmentation, brainstorming).
\end{itemize}
This complementarity mirrors the broader machine learning landscape, where autoregressive and diffusion approaches coexist in image generation, speech synthesis, and music generation.

\subsection{Limitations}
\label{sec:limitations}

\begin{enumerate}[leftmargin=*,itemsep=2pt]
  \item \textbf{Small scale.}
  Our models are 123--163M parameters trained on 50M tokens.
  At 1B+ scale, the dynamics may differ: AR overfitting may be less pronounced with more data, and MDLM may benefit from scale differently.

  \item \textbf{Parameter mismatch.}
  The MDLM has 31.6\% more parameters (Section~\ref{sec:param-diff}).
  This is an inherent asymmetry, but it complicates attribution of observed differences.

  \item \textbf{Limited diversity metrics.}
  We report Distinct-$n$, Self-BLEU, and opening uniqueness over 1{,}000 samples. Additional metrics (MAUVE, coherence scoring, human evaluation) would further strengthen the analysis.

  \item \textbf{Single dataset.}
  TinyStories is a narrow domain (children's stories).
  Generalization to code, scientific text, dialogue, and multilingual data is unknown.

  \item \textbf{MDLM still converging.}
  The MDLM had not plateaued at 20K steps.
  Extended training could change the relative quality of generated text.

  \item \textbf{Single seed.}
  All experiments use one random seed.
  Multiple seeds with confidence intervals would strengthen all claims.

  \item \textbf{No downstream evaluation.}
  We do not evaluate on classification, QA, or other downstream tasks.

  \item \textbf{Sampler not optimized.}
  The MDLM sampler (confidence-based unmasking, temperature annealing, repetition penalty) was not extensively tuned.
  Better sampling strategies (e.g., nucleus-based unmasking, classifier-free guidance for language) could improve MDLM output quality.
\end{enumerate}

% ── 6. Conclusion ──────────────────────────────────────────────────────
\section{Conclusion}

We present a controlled comparison between autoregressive and masked diffusion language models, trained on identical data, budget, and hardware.
Our findings:

\begin{enumerate}[leftmargin=*,itemsep=2pt]
  \item \textbf{Training throughput is near-identical}: MDLM trains at 95.5\% of AR throughput, suggesting that diffusion training overhead is minimal at the Transformer scale.

  \item \textbf{Convergence regimes differ}: AR converges in ${\sim}$14K steps then overfits; MDLM converges monotonically through 20K steps with no plateau, implying different optimal training lengths.

  \item \textbf{Generation diversity differs quantitatively}: Over 1{,}000 samples, AR produces fluent but structurally repetitive text (99.8\% same first word, 3.3\% unique openings), while MDLM produces diverse text (36.1\% unique first words, 93.4\% unique openings) with higher Distinct-$n$ and lower Self-BLEU, confirming a diversity--fluency trade-off that implies the paradigms are complementary rather than competing.
\end{enumerate}

We release all code, trained models, and data pipelines at \url{https://github.com/caiovicentino/arche}.
Future work should (i) validate the diversity observation quantitatively, (ii) extend the comparison to larger scales and diverse datasets, and (iii) train MDLM for longer to determine its convergence characteristics.

% ── References ─────────────────────────────────────────────────────────
\bibliographystyle{plainnat}

% ── Appendix ───────────────────────────────────────────────────────────
\appendix

\section{MDLM Implementation Details}
\label{app:mdlm}

\paragraph{Cosine noise schedule.}
The masking probability follows $\gamma(t) = 1 - \cos^2(\pi t / 2)$ for $t \in [0, 1]$.
At $t = 0$, no tokens are masked (clean input); at $t = 1$, all tokens are masked (pure noise).
During training, $t$ is sampled uniformly for each example in the batch.

\paragraph{Confidence-based unmasking.}
During generation, at each of the $S = 100$ denoising steps, the model produces logits for all masked positions.
We compute the maximum softmax probability at each masked position and unmask the top-$k$ most confident positions, where $k = \lfloor L / S \rfloor$ and $L$ is the sequence length.
Ties are broken randomly.

\paragraph{Temperature annealing.}
We apply a linear temperature schedule from $\tau_{\text{start}} = 1.2$ to $\tau_{\text{end}} = 0.5$ across the $S$ denoising steps.
High temperature early in generation encourages diverse initial predictions; low temperature late in generation sharpens the distribution for coherent refinement.

\paragraph{Repetition penalty.}
We apply a frequency-based repetition penalty: for any token that has already been unmasked, we divide its logit by a penalty factor of 1.3.
This reduces the tendency to produce degenerate repetitive sequences.

\section{Compute Resources}
\label{app:compute}

\begin{table}[H]
\centering
\caption{Compute budget for all experiments reported in this paper.}
\smallskip
\begin{tabular}{@{}llcc@{}}
\toprule
Model & Hardware & Time & Cost \\
\midrule
AR (123.6M, 20K steps) & H100 80GB (Colab) & 107.9 min & ${\sim}$\$35 \\
MDLM (162.7M, 20K steps) & H100 80GB (Colab) & 113.0 min & ${\sim}$\$35 \\
\midrule
\textbf{Total} & & \textbf{220.9 min} & $\boldsymbol{\sim}$\textbf{\$70} \\
\bottomrule
\end{tabular}
\end{table}

\section{Data}
\label{app:data}

All experiments use 50M tokens from TinyStories \citep{tinystories}, a dataset of short children's stories generated by GPT-3.5 and GPT-4, specifically designed for training and evaluating small language models.
The dataset was tokenized using the GPT-2 tokenizer (vocabulary size 50{,}257).
We split the tokenized data into training and validation sets, using the last 5\% of chunks as the validation set.

\end{document}